%% file: neurips_2026.tex
\pdfoutput=1
\documentclass{article}

\usepackage[main, final]{neurips_2026}

\usepackage[utf8]{inputenc} 
\usepackage[T1]{fontenc}    
\usepackage{hyperref}       
\usepackage{url}            
\usepackage{booktabs}       
\usepackage{amsfonts}       
\usepackage{nicefrac}       
\usepackage{microtype}      
\usepackage{xcolor}         
\newcommand{\EVAN}{\rev{\texttt{EVAN}}}
\usepackage{bm}
\usepackage{graphicx}
\usepackage[table]{xcolor}
\colorlet{revcolor}{black}
\newcommand{\rev}[1]{{\color{revcolor}#1}}
\usepackage{enumitem}
\usepackage{arydshln}
\usepackage{arydshln}
\usepackage{amsmath}
\usepackage{amssymb}
\usepackage{amsthm}
\usepackage{algorithm}
\usepackage{algpseudocode}
\usepackage{bbm}

\title{Grounding Multimodal Reasoning with Evidence-Ablated Negatives}

\author{%
  Chenxi Liu \\
  Department of Computer Science\\
  University of Maryland, College Park\\
  College Park, MD 20742 \\
  \texttt{cxliu539@umd.edu} \\
  \And
  Tianyi Xiong \\
  Department of Computer Science\\
  University of Maryland, College Park\\
  College Park, MD 20742 \\
  \texttt{txiong23@umd.edu} \\
  \And
  Yanshuo Chen \\
  Department of Computer Science\\
  University of Maryland, College Park\\
  College Park, MD 20742 \\
  \texttt{cys@umd.edu} \\
  \And
  Heng Huang \\
  Department of Computer Science\\
  University of Maryland, College Park\\
  College Park, MD 20742 \\
  \texttt{heng@umd.edu} \\
}

\begin{document}

\maketitle

\begin{abstract}
Reinforcement learning with verifiable rewards (RLVR) has become a powerful tool for improving multimodal reasoning. 
However, standard outcome supervision suffers from credit assignment ambiguity, as a model may receive positive rewards by exploiting language priors or dataset shortcuts rather than grounding its reasoning in the requisite visual evidence.
This makes high-quality negative responses critical, since effective negatives should remain plausible under the current policy while exposing the shortcut that should be suppressed. 
In this paper, we propose \EVAN{}, a multimodal reinforcement learning framework utilizing \emph{\rev{EVidence-Ablated Negatives}} to train LMMs against their own language-prior failures. 
\EVAN{} first identifies highly visually dependent examples using a scaled image information gain metric and preserves data diversity with a mixed-ranking strategy. 
It then constructs evidence-ablated images by masking the most discriminative image patches, eliciting plausible but visually ungrounded responses from the current policy. 
During policy optimization, these negative responses are incorporated as bounded auxiliary signals under the clean image, penalizing shortcut learning while maintaining RL stability. 
Experiments across challenging multimodal reasoning benchmarks demonstrate that \EVAN{} mitigates modality imbalance and improves over strong RLVR baselines.
\rev{Our code is available at \url{https://github.com/DawnLIU35/EVAN.git}.}
\end{abstract} 

\input{sec/intro}
\input{sec/related_works}
\input{sec/preliminaries}
\input{sec/methodology}
\input{sec/experiments}

\section{Conclusion}

We presented \EVAN{}, a multimodal RL framework that improves visual reasoning by learning from evidence-ablated negatives.
\EVAN{} selects visually dependent examples with scaled-IIG, constructs evidence-ablated images by removing query-critical visual evidence, and incorporates the resulting negatives into clean-image RLVR with stability controls.
Experiments on challenging multimodal reasoning benchmarks show that \EVAN{} consistently improves over strong RLVR baselines across model scales.
\rev{Ablations show that these gains come from localizing query-critical evidence rather than from extra rollouts or the masked area alone, and that bounded negative advantages and rollout filtering are both needed for stable training.
The gains also extend to Qwen3-VL-8B and transfer beyond structured reasoning to chart understanding, general visual question answering, and counterfactual visual grounding, with no added parameters or computation at inference time.}
These results suggest that targeted evidence ablation provides useful auxiliary supervision for reducing language-prior shortcuts in LMMs.

\bibliographystyle{plainnat}
\bibliography{ref}

\appendix
\input{sec/appendix}

\end{document}

%% file: sec/intro.tex
\section{Introduction}
\label{sec:intro}
\emph{Large Multimodal Models} (LMMs) have made substantial progress by coupling vision encoders with pre-trained \emph{Large Language Models} (LLMs), enabling strong performance on diverse visual understanding and reasoning tasks~[\citealp{liu2023visual,liu2024llavanext,Qwen2.5-VL,tong2024cambrian,chen2024internvl,2023GPT4VisionSC,xiong2024llavacritic,chen2023gpt,chen2024your}, \rev{\citealp{xiong2026multi,xiong2026rubrics}}].
More recently, \emph{Reinforcement Learning with Verifiable Rewards} (RLVR), often instantiated with \emph{Group Relative Policy Optimization} (GRPO)~\citep{shao2024deepseekmath}, has emerged as a promising post-training paradigm for improving \rev{model} reasoning through online rollouts and task-level correctness signals~[\citealp{guo2025deepseek,chen2025r1v,shen2025vlm,zheng2025easyr1}, \rev{\citealp{liu2026learning,liu2025explore,jia2026blackbox}}].
However, verifiable rewards typically assess only whether the final answer is correct, without indicating whether a correct prediction is genuinely grounded in the visual input or instead driven by language priors, dataset shortcuts, or spurious question–answer correlations.
As a result, LMMs can still suffer from \emph{modality imbalance}, where the language backbone dominates the reasoning process while visual evidence is underutilized~\citep{liu2024paying,jiang2024modality}. 
This failure mode is especially problematic for visual reasoning, since a model may obtain correct rewards during training yet remain weakly grounded in the input image, leading to brittle perception and hallucinated responses~\citep{yu2024rlhf,zhao2023beyond}.

A natural way to address modality imbalance is to intervene on the visual input and test whether the model's behavior changes accordingly. 
Prior work perturbs, masks, or contrasts images to reduce language bias in visual question answering~\citep{niu2021counterfactual,liang2020learning}, and recent multimodal RL methods use perturbed images to estimate visual dependency or adjust policy updates~\citep{wang2025perception,huang2025spotlight,liu2025noisyrollout}. 
These studies highlight the value of visual interventions, but they do not necessarily produce informative negative samples for RLVR. 
Most interventions are still coarse and task-agnostic. 
They perturb images without first selecting examples whose solutions strongly require visual evidence, and the perturbations are not optimized to expose the model's most plausible language-prior failure for the given query. 
Consequently, the induced negatives can be too easy, out of distribution, or only weakly connected to the shortcut that causes modality imbalance. 
This limitation is important because recent analyses of RLVR show that penalizing incorrect responses can itself provide a strong learning signal, and that plausible negative trajectories are more useful than arbitrary failures~\citep{zhu2025surprising, di2026not}. 
These findings shift the focus from simply perturbing images to constructing high-quality negatives. 
This motivates a targeted approach that first identifies visually dependent examples and then ablates query-critical visual evidence to elicit hard, plausible, and visually ungrounded negative responses for policy optimization.

In this paper, we propose \EVAN{}, a multimodal reinforcement learning framework that learns from \emph{\rev{EVidence-Ablated Negatives}} to encourage LMMs to ground their reasoning in query-critical visual evidence. 
The key idea is to expose the model’s language-prior shortcuts only on examples where visual evidence is truly necessary. 
To this end, \EVAN{} first estimates \emph{scaled image information gain} (\rev{scaled-IIG}) for each training example and applies a mixed-ranking strategy to select visually dependent samples while preserving data diversity. 
For each selected sample, \EVAN{} identifies the image patches that most support the verified response and constructs an evidence-ablated image by masking these query-critical patches. 
The evidence-ablated image is then used to sample plausible but visually ungrounded responses from the current policy, which serve as hard negative responses. 
During policy optimization, \EVAN{} incorporates these online evidence-ablated negatives together with clean-image rollouts. 
The evidence-ablated negatives are optimized under the clean image, so the policy learns to suppress language-prior failures when the relevant visual evidence is available. 
To maintain RL stability, \EVAN{} filters out low-probability or overly long evidence-ablated responses and assigns bounded negative advantages, preventing uninformative negatives from dominating policy updates. 
Experiments on challenging visual reasoning benchmarks show that \EVAN{} improves multimodal reasoning performance and yields more visually grounded behavior.

Overall, \textbf{our contributions} can be summarized as follows:
\begin{itemize}[leftmargin=2em]
    \setlength{\itemsep}{0.3em}

    \item We propose \EVAN{}, a multimodal reinforcement learning framework that identifies visually dependent examples with scaled-IIG and constructs evidence-ablated images by masking query-critical patches. These images elicit plausible but visually ungrounded responses from the current policy.

    \item We introduce a stable training procedure for \EVAN{} that incorporates online evidence-ablated negatives into RLVR. The negatives are filtered for clean-image plausibility and optimized under the clean image with bounded negative advantages.

    \item We evaluate \EVAN{} on challenging multimodal reasoning benchmarks and show that it improves over strong RLVR baselines. Ablations and further analyses validate the roles of visually dependent example selection, evidence-ablated image construction, and stable negative optimization.
\end{itemize}

%% file: sec/related_works.tex
\section{Related Work}
\label{related_work}

\textbf{RLVR for Multimodal Reasoning.}
Reinforcement learning has become a central post-training tool for improving reasoning and alignment in large language models~\citep{ouyang2022training,jaech2024openai,guo2025deepseek,team2025kimi}. 
Recent work extends \emph{Reinforcement Learning with Verifiable Rewards} (RLVR), often instantiated with GRPO~\citep{shao2024deepseekmath}, to multimodal reasoning tasks such as visual math reasoning and academic question answering~\citep{chen2025r1v,shen2025vlm,zheng2025easyr1,meng2025mm,huang2025vision,peng2025lmm,yang2025r1,yu2025perception,zhan2025vision,liu2025seg}. 
These methods mainly optimize final-answer correctness from online rollouts. 
However, outcome rewards do not reveal whether a correct answer is grounded in visual evidence or produced from language priors. 
Recent studies further show that negative responses can provide important RLVR signals, especially when they are plausible rather than arbitrary failures~\citep{zhu2025surprising, di2026not}. 
\EVAN{} is complementary to standard multimodal RLVR. 
It preserves clean-image policy optimization while introducing online evidence-ablated negatives that expose modality imbalance.

\textbf{Visual Interventions for Modality Imbalance.}
Large Multimodal Models often inherit strong linguistic priors from pre-trained language backbones~\citep{liu2023visual,liu2024llavanext,Qwen2.5-VL,tong2024cambrian,chen2024internvl}. 
This can lead to \emph{modality imbalance}, where the model answers from language priors or dataset shortcuts instead of visual evidence~\citep{liu2024paying,jiang2024modality, xiong2026phycritic}. 
Prior work studies this issue through hallucination analysis, counterfactual VQA, and visual grounding evaluation~\citep{zhao2023beyond,yu2024rlhf,niu2021counterfactual}. 
Other methods intervene on images by masking, corrupting, cropping, or contrasting visual inputs to reduce language bias or construct dispreferred responses~\citep{liang2020learning,pi2024strengthening,zhou2024aligning,wang2024mdpo}. 
Recent perception-aware multimodal RL methods also use perturbed images to regularize policy updates~\citep{wang2025perception,huang2025spotlight,liu2025noisyrollout,liu2025modality} \rev{or impose KL constraints between original and code-edited chart images~\citep{zhang2026see}}. However, these signals may not effectively reflect the language-prior failures of the online policy alone training\rev{, or are limited to specific data such as charts}.
In contrast, \EVAN{} first selects visually dependent examples with \rev{scaled-IIG} and then constructs robust evidence-ablated images by masking query-critical patches. 
The resulting images elicit hard negative responses from the current policy, which are filtered and optimized under the clean image with bounded negative advantages.

%% file: sec/preliminaries.tex
\section{Preliminaries}
\label{preliminaries}

\paragraph{Reinforcement Learning.}
 Group Relative Policy Optimization (GRPO)~\citep{shao2024deepseekmath,guo2025deepseek} estimates advantages from a group of sampled responses without training an additional critic. 
Given a query $q$, the behavior policy $\pi_{\theta_\text{old}}$ generates a group of $G$ candidate responses $\{o_i\}_{i=1}^G$. 
Each response receives a reward \rev{$R_i$}, and its advantage is computed by centering and normalizing the rewards within the group:
\begin{equation}
\label{eq:advantage}
\rev{\hat{A}_{i}} = \frac{\rev{R_i} - \text{mean}(\{\rev{R_i}\}_{i=1}^G)}{\text{std}(\{\rev{R_i}\}_{i=1}^G)}.
\end{equation}
\rev{All tokens of $o_i$ share this response-level advantage.}
We further adopt Decoupled Clip and Dynamic sAmpling Policy Optimization (DAPO)~\citep{yu2025dapo} for stable long-form reasoning optimization. 
DAPO uses asymmetric clipping to control policy updates and applies dynamic sampling to keep only groups that contain both correct and incorrect responses. 
For readability, we omit the image input in the following preliminary objective and restore it in the multimodal formulation of \EVAN{}. 
The resulting token-level objective is:
{\small
\begin{equation}
\begin{aligned}
\mathcal{J}_{\text{DAPO}}(\theta) =\quad& \mathbb{E}_{(q,o^*)\sim \mathcal{Q}, \{o_i\}_{i=1}^G\sim \pi_{\theta_\text{old}}(\cdot\mid q)}
\Bigg[\frac{1}{\sum_{i=1}^{G}|o_i|}\sum_{i=1}^{G}\sum_{t=1}^{|o_i|} \\
&\hspace{-4em} \min \Big( r_{i,t}(\theta) \rev{\hat{A}_{i}},  
\ \text{clip} \Big( r_{i,t}(\theta), 1 - {\varepsilon_{\text{low}}}, 1 + {\varepsilon_{\text{high}}} \Big) \rev{\hat{A}_{i}} \Big) \Bigg]
\\
\text{s.t.}\quad& 0< \Big|\{o_i\mid\texttt{is\_equivalent}(o^*,o_i)\}\Big|< G,
\label{eq:dapoloss}
\end{aligned}
\end{equation}
}
where $r_{i,t}(\theta)=\frac{\pi_{\theta}(o_{i,t} \mid q, o_{i,<t})}{\pi_{\theta_{\text{old}}}(o_{i,t} \mid q,o_{i,<t})}$ denotes the token-level importance ratio, and $o^*$ is the ground-truth answer used for reward computation.

%% file: sec/methodology.tex
\section{Methodology}
\label{methodology}

\EVAN{} is designed to mitigate modality imbalance by training LMMs against hard negative responses that reveal language-prior shortcuts. 
The framework contains two stages. 
First, we mine visually dependent examples and construct evidence-ablated images by masking query-critical visual patches. 
These images selectively weaken the evidence needed to support the verified response, prompting the current policy to generate plausible but visually ungrounded responses. 
Second, we incorporate the resulting online evidence-ablated negatives into GRPO-based policy optimization with stability controls. 
The negatives are filtered and optimized under the clean image with bounded negative advantages, allowing the model to suppress shortcut-driven responses while preserving standard clean-image rollout optimization. 
The full procedure of \EVAN{} is summarized in Algorithm~\ref{alg:earn}.

\begin{algorithm}[t]
\caption{\EVAN{}: Learning from \rev{EVidence-Ablated Negatives}}
\label{alg:earn}
\small
\begin{algorithmic}[1]
\Require Training set $\mathcal{D}=\{(q_j,I_j,o_j^*)\}_{j=1}^{N}$, policy $\pi_\theta$, selection ratio $\tau$, mask ratio $\rho$
\Ensure Updated policy $\pi_\theta$

\State \textbf{Stage 1: Evidence-ablated image construction}
\For{each example $(q_j,I_j,o_j^*)\in\mathcal{D}$}
    \State Compute the visual-dependency score $s_j=\mathrm{sIIG}(o_j^*,q_j,I_j)$.
\EndFor
\State Select a visually dependent subset $\mathcal{S}_{\mathrm{adv}}$ with RankMix using ratio $\tau$.
\For{each selected example $j\in\mathcal{S}_{\mathrm{adv}}$}
    \State Compute patch scores from the absolute gradient of $\mathrm{sIIG}(o_j^*,q_j,I_j)$ with respect to image patches.
    \State Mask the top $\rho$ fraction of patches to construct the evidence-ablated image $I_j^{\mathrm{adv}}$.
\EndFor
\State Form $\mathcal{D}_{\mathrm{train}}=\mathcal{D}\cup\{(q_j,I_j,I_j^{\mathrm{adv}},o_j^*)\mid j\in\mathcal{S}_{\mathrm{adv}}\}$.

\Statex
\State \textbf{Stage 2: Stable policy optimization}
\For{each training step}
    \State Set $\pi_{\theta_{\mathrm{old}}}\leftarrow \pi_\theta$ and sample a mini-batch $\mathcal{B}\subseteq\mathcal{D}_{\mathrm{train}}$.
    \For{each example $j\in\mathcal{B}$}
        \State Generate clean rollouts $\mathcal{O}_j^c=\{o_{j,i}^c\}_{i=1}^{G}\sim\pi_{\theta_{\mathrm{old}}}(\cdot\mid q_j,I_j)$.
        \State Compute rewards and DAPO advantages $\{\hat A_{j,i}^c\}_{i=1}^{G}$ for the clean rollout group.
        \State Initialize the clean-image update set $\mathcal{U}_j\leftarrow\{(o_{j,i}^c,\hat A_{j,i}^c,I_j)\}_{i=1}^{G}$.
        \If{an evidence-ablated image $I_j^{\mathrm{adv}}$ is available}
            \State Generate evidence-ablated rollouts $\widetilde{\mathcal{O}}_j^a=\{\tilde o_{j,k}^a\}_{k=1}^{M}\sim\pi_{\theta_{\mathrm{old}}}(\cdot\mid q_j,I_j^{\mathrm{adv}})$.
            \State \rev{Randomly select one incorrect evidence-ablated rollout as the candidate $o_j^a$, if any exists.}
            \If{\rev{$o_j^a$ passes the clean-image likelihood and response length filters}}
                \State \rev{Assign $o_j^a$ the bounded negative advantage} $\hat A_j^a=-\min_i|\hat A_{j,i}^c|$.
                \State Add it to the clean-image update set: $\mathcal{U}_j\leftarrow\mathcal{U}_j\cup\{(o_j^a,\hat A_j^a,I_j)\}$.
            \EndIf
        \EndIf
    \EndFor
    \State Update $\pi_\theta$ on \rev{$\mathcal{U}=\bigcup_{j\in\mathcal{B}}\mathcal{U}_j$} with the DAPO objective\rev{.}
\EndFor
\end{algorithmic}
\end{algorithm}

\subsection{Evidence-ablated Data Mining}
\label{sec:adv_mining}

The goal of evidence-ablated data mining is to identify examples whose verified responses require visual evidence and construct evidence-ablated images that expose language-prior failures. 
Applying visual interventions uniformly to all samples is inefficient. 
Different examples require different degrees of visual information, and perturbing weakly visual-dependent samples may provide little useful training signal. 
Meanwhile, overly destructive interventions may simply make the image uninformative and produce off-distribution failures. 
\EVAN{} therefore follows a targeted procedure. 
It first estimates the visual dependency of each example, then selects a category-balanced subset with high dependency, and finally masks query-critical patches that suppress the same dependency signal under the original query.

\paragraph{Scaled image information gain.}
We use scaled image information gain (\rev{scaled-IIG})\rev{, denoted by $\mathrm{sIIG}$ in equations,} to measure how much a response relies on the image. 
Let $\mathcal{D}=\{(q_j,I_j,o_j)\}_{j=1}^{N}$ be the candidate pool, where $q_j$ is the query, $I_j$ is the image, and $o_j$ is the response used for scoring. 
For a response $o$, we first compute its average log-likelihood under image $I$:
\begin{equation}
\label{eq:avg_log_likelihood}
\ell_{\theta}(o;q,I)
=
\frac{1}{|o|}
\sum_{t=1}^{|o|}
\log \pi_{\theta}(o_t \mid q,I,o_{<t}).
\end{equation}
We then define the visual-dependency score of example $j$ as the scaled likelihood difference between the original image and a blank image $I_{\emptyset}$:
\begin{equation}
\label{eq:visual_dependency_score}
s_j
=
\mathrm{sIIG}(o_j,q_j,I_j)
=
\frac{
\ell_{\theta}(o_j;q_j,I_j)
-
\ell_{\theta}(o_j;q_j,I_{\emptyset})
}{
-\ell_{\theta}(o_j;q_j,I_{\emptyset})+\eta
}.
\end{equation}
Here $\eta$ is a small numerical stabilizer. 
A larger $s_j$ means that replacing the image with a blank input causes a larger relative drop in response likelihood. 
Thus, a high score indicates that the response is supported by visual evidence rather than mainly by language priors.
This score is model-centered: it measures how much the LMM actually relies on the image, rather than assuming visual dependency from the task category or dataset annotation.

\paragraph{Balanced selection with RankMix.}
Selecting only the globally highest scaled-IIG examples can reduce data diversity. 
The top-ranked samples may concentrate in a few categories where blank-image likelihood drops are especially large. 
To preserve both visual dependency and category coverage, we use RankMix, which combines global and category-wise rankings.

Let $c_j$ denote the category of example $j$, let $\mathcal{I}=\{1,\ldots,N\}$ be the full index set, and let $\mathcal{I}_c=\{j:c_j=c\}$ be the index set for category $c$. 
For any index set $\mathcal{A}$, let $\operatorname{Top}_{\tau}(\mathcal{A};s)$ return the top $\tau$ fraction of examples in $\mathcal{A}$ ranked by scaled-IIG score $s$. 
RankMix first forms a core set of examples that are strong under both global and category-wise rankings:
\begin{equation}
\label{eq:rankmix_core}
\mathcal{C}
=
\left\{
j\in\mathcal{I}
\mid
j\in\operatorname{Top}_{\tau}(\mathcal{I};s)
\ \mathrm{and}\
j\in\operatorname{Top}_{\tau}(\mathcal{I}_{c_j};s)
\right\}.
\end{equation}
This criterion keeps examples that are globally high-dependency while also being competitive within their own categories. 
If the core set is smaller than the target size, RankMix fills the remaining slots using a mixed rank score. 
Let $r(j;\mathcal{A})$ be the descending scaled-IIG rank of example $j$ within an index set $\mathcal{A}$, with smaller ranks indicating larger scores. 
The fill score is defined as
\begin{equation}
\label{eq:rankmix_fill}
m_j
=
\left(
1-\frac{r(j;\mathcal{I})}{|\mathcal{I}|}
\right)
\left(
1-\frac{r(j;\mathcal{I}_{c_j})}{|\mathcal{I}_{c_j}|}
\right).
\end{equation}
A larger $m_j$ indicates that the example ranks highly both globally and within its category. 
The final selected set is formed by $\mathcal{C}$ plus the highest-scoring remaining examples according to $m_j$ until the target size is reached. 
\rev{Since $m_j$ ranks every example outside $\mathcal{C}$, a category with no example in $\mathcal{C}$ can still contribute examples that rank highly within that category.}
In this way, RankMix keeps the mining pool focused on high visual dependency while avoiding diversity collapse.

\paragraph{Scaled-IIG-guided adversarial masking.}
After selecting visually dependent examples, we construct evidence-ablated images by masking the regions that most affect the same scaled-IIG signal used for selection. 
The goal is not to make the whole image unreadable. 
Instead, the intervention should remove high-influence visual evidence in a localized patch-level manner while keeping the textual query unchanged. 
For a selected example, let $o_j^*$ denote a verified correct target response. 
We first compute how sensitive the scaled-IIG score is to each visual patch. 
Let \rev{$\mathcal{P}_j$ be the set of non-overlapping patches that partition the image, and let $P_{j,r}\in\mathcal{P}_j$ denote the $r$-th patch}. 
For each patch, we aggregate the absolute scaled-IIG gradient:
\begin{equation}
\label{eq:iig_patch_score}
a_{j,r}
=
\sum_{u\in P_{j,r}}
\left|
\frac{\partial\,\mathrm{sIIG}(o_j^*,q_j,I_j)}
{\partial I_j(u)}
\right|.
\end{equation}
Here, $u$ indexes a pixel location in $I_j$, and $I_j(u)$ denotes the corresponding pixel value. We then select the top $\rho$ fraction of patches by $a_{j,r}$ and set their pixels to black:
\begin{equation}
\label{eq:iig_patch_mask}
I_j^{\mathrm{adv}}(u)
=
\begin{cases}
0, & u\in \bigcup_{P\in\mathcal{M}_j} P,\\
I_j(u), & \mathrm{otherwise},
\end{cases}
\quad
\mathcal{M}_j=\operatorname{Top}_{\rho}(\mathcal{P}_j;a).
\end{equation}
This masking operation aligns the image intervention with the mining criterion. 
Rather than applying generic corruption or relying on fragile pixel-level perturbations, it uses \rev{scaled-IIG} to localize the visual evidence most responsible for a valid target response and removes that evidence explicitly. 
Because the intervention operates at the patch level, it provides an evidence ablation while still targeting the evidence used by the current model. 
Conditioned on $I_j^{\mathrm{adv}}$, the model is more likely to fall back on high-probability textual patterns and dataset priors under the same query.

We store the evidence-ablated image together with its corresponding clean example:
\begin{equation}
\label{eq:adv_image_pool}
\mathcal{D}_{\mathrm{adv}}
=
\{(q_j,I_j,I_j^{\mathrm{adv}})\mid j\in\mathcal{S}_{\mathrm{adv}}\},
\end{equation}
where $\mathcal{S}_{\mathrm{adv}}$ denotes the RankMix-selected index set. 
The prepared evidence-ablated entries are appended to the original training set and later used to sample online evidence-ablated negatives:
\begin{equation}
\label{eq:augmented_dataset}
\mathcal{D}_{\mathrm{train}}
=
\mathcal{D}
\cup
\mathcal{D}_{\mathrm{adv}}.
\end{equation}

\subsection{Stable Policy Optimization}
\label{sec:stable_policy_optimization}

Evidence-ablated images are used to elicit hard negative responses that reveal language-prior shortcuts. 
However, these responses are sampled under weakened visual evidence, so directly treating them as ordinary clean-image on-policy rollouts can introduce unstable gradients. 
\EVAN{} addresses this by using evidence-ablated rollouts as bounded auxiliary signals. 
The clean-image rollout group follows the standard DAPO update, while examples with prepared evidence-ablated images receive an additional evidence-ablated branch. 
Only one filtered incorrect response from this branch is appended to the actor update.

\paragraph{Evidence-ablated image rollout.}
For a selected example $(q_j,I_j,o_j^*)$ with evidence-ablated image $I_j^{\mathrm{adv}}$, \EVAN{} samples an auxiliary rollout group under the evidence-ablated visual condition:
\begin{equation}
\label{eq:adv_rollout_group}
\widetilde{\mathcal{O}}_j^{a}
=
\{\tilde o_{j,k}^{a}\}_{k=1}^{M},
\quad
\tilde o_{j,k}^{a}
\sim
\pi_{\theta_{\mathrm{old}}}(\cdot\mid q_j,I_j^{\mathrm{adv}}).
\end{equation}
We retain only evidence-ablated rollouts that have incorrect final answers:
\begin{equation}
\label{eq:incorrect_adv_set}
\mathcal{O}_j^{a}
=
\left\{
\tilde o_{j,k}^{a}\in\widetilde{\mathcal{O}}_j^{a}
\mid
\texttt{is\_equivalent}(o_j^*,\tilde o_{j,k}^{a})=0
\right\}.
\end{equation}
Since the visual evidence supporting the verified response has been weakened, these incorrect responses serve as candidate language-prior failures under the same query.

\paragraph{Rollout filtering.}
A wrong response generated under the evidence-ablated image is useful only if it remains plausible under the clean visual context.
If it is far outside the clean-image policy distribution, penalizing it may introduce noisy gradients rather than correct a realistic shortcut. 
We therefore filter evidence-ablated candidates by comparing their clean-image likelihood with the clean rollout group. 
Let $\mathcal{O}_j^{c}$ denote the clean rollout group, and let $\ell_{\theta_{\mathrm{old}}}(o;q,I)$ denote the token-mean log probability defined above. 
\rev{The evidence-ablated candidates that pass the filter are}
\begin{equation}
\label{eq:logp_filter}
\mathcal{O}_{j,\mathrm{keep}}^{a}
=
\left\{
o^a\in\mathcal{O}_j^{a}
\mid
\ell_{\theta_{\mathrm{old}}}(o^a;q_j,I_j)
>
\min_{o^c\in\mathcal{O}_j^{c}}
\ell_{\theta_{\mathrm{old}}}(o^c;q_j,I_j),
\quad
|o^a|
\leq
\max_{o^c\in\mathcal{O}_j^{c}} |o^c|
\right\}.
\end{equation}
This relative criterion keeps evidence-ablated negatives that are more likely than the weakest clean rollout under the clean image. 
The length constraint removes evidence-ablated negatives that are longer than all clean rollouts for the same prompt, keeping appended negatives comparable to the clean rollout group. 
\rev{For each prompt with a non-empty $\mathcal{O}_j^{a}$, we randomly select one candidate $o_j^{a}$ from $\mathcal{O}_j^{a}$ and append it only if $o_j^{a}\in\mathcal{O}_{j,\mathrm{keep}}^{a}$.
Otherwise, no evidence-ablated negative is appended, and no other candidate is tried.}
This ensures that each training example contributes at most one online evidence-ablated negative, keeping the auxiliary signal bounded.

\paragraph{Bounded negative advantage.}
The selected evidence-ablated negative response is appended to the actor batch, but it does not change the clean group statistics. 
Let $\{\hat A_{j,i}^{c}\}_{i=1}^{G}$ be the DAPO advantages of the clean rollout group. 
We assign the evidence-ablated negative a bounded negative advantage:
\begin{equation}
\label{eq:adv_advantage}
\hat A_j^{a}
=
-\min_{1\leq i\leq G}
\left|
\hat A_{j,i}^{c}
\right|.
\end{equation}
This design uses the clean rollout group as a proxy for the current sample's uncertainty. 
Higher outcome variance among clean rollouts indicates greater sample-level uncertainty and warrants a stronger auxiliary signal, whereas low outcome variance suggests that the model's behavior on this sample is already stable, so \EVAN{} applies only a mild correction to avoid dominating the update.
\rev{Appendix~\ref{app:bounded_adv} illustrates this behavior under binary rewards.}

The retained evidence-ablated negative is appended to the clean rollout group and optimized in the same actor update. 
Although $o_j^{a}$ is generated from the evidence-ablated image $I_j^{\mathrm{adv}}$, it is paired with the clean image $I_j$ during optimization, which is the same visual condition used for clean rollouts. 
Its importance ratio is therefore computed as
\begin{equation}
\label{eq:adv_mbpo_ratio}
\rev{r_{j,t}^{a}(\theta)}
=
\frac{
\pi_{\theta}(o_{j,t}^{a}\mid q_j,I_j, o_{j,<t}^{a})
}{
\pi_{\theta_{\mathrm{old}}}(o_{j,t}^{a}\mid q_j,I_j, o_{j,<t}^{a})
}.
\end{equation}
Clean-image responses use their original DAPO advantages, while the appended evidence-ablated negative uses the bounded negative advantage $\hat A_j^{a}$. 
This formulation penalizes a plausible but visually ungrounded response under the clean visual context, where the relevant evidence is available. 
It preserves standard clean-image policy optimization while adding a bounded response-level signal against shortcut learning.

%% file: sec/experiments.tex
\section{Experiments}
\label{experiments}

In this section, we first introduce the implementation details, including training details, datasets, evaluation protocol and baseline methods. Subsequently, we present our main results comparing \EVAN{} with baseline methods on several visual reasoning benchmarks, demonstrating the effectiveness of \EVAN{}. In addition, the ablation study provides a closer look at \EVAN{} and verifies the contributions of its individual components. Lastly, we include additional experimental results for further analysis.

\subsection{Implementation Details}
\label{sec:implementation}

\textbf{Training details.}
Following recent studies that apply GRPO to train LMMs~\citep{wang2025perception,huang2025spotlight}, we adopt \texttt{Qwen2.5-VL-3B-Instruct}\rev{,} \texttt{Qwen2.5-VL-7B-Instruct}~\citep{Qwen2.5-VL}\rev{, and \texttt{Qwen3-VL-8B-Instruct}~\citep{bai2025qwen3}} as backbone models and train them on the ViRL39K dataset~\citep{wang2025vl}. We train each model for 2 epochs with a learning rate of $1\times10^{-6}$ and a prompt-generation batch size of 256, generating $8$ responses per prompt. We set the maximum response length to 4096 for both the 3B and 7B models\rev{, and to 8192 for the 8B model}. To construct evidence-ablated images, we select $\tau=20\%$ of the training set and mask $50\%$ of the most influential patches using \rev{$16\times16$} patches. All experiments are conducted using the \texttt{verl} framework~\citep{sheng2024hybridflow}. Additional details are provided in the Appendix.

\noindent\textbf{Baselines:} We compare our models against a comprehensive set of state-of-the-art open-source reasoning LMMs with publicly released model weights, including VL-Rethinker~\citep{wang2025vl}, R1-ShareVL~\citep{yao2025r1}, NoisyRollout~\citep{liu2025noisyrollout}, PAPO~\citep{wang2025perception}, \rev{BiPS~\citep{zhang2026see},} and VPPO~\citep{huang2025spotlight}. We also train the base model with DAPO and use it as our baseline.

\noindent\textbf{Evaluation benchmark:} Following recent multimodal RL studies~\citep{wang2025perception}, our evaluation benchmark covers mathematical and geometric reasoning, including Geometry3K~\citep{lu2021inter}, MathVista~\citep{lu2023mathvista}, MathVerse~\citep{zhang2024mathverse}, We-Math~\citep{qiao2025we}, DynaMath~\citep{zou2024dynamath}, MathVision~\citep{wang2024measuring}, and MMK12~\citep{meng2025mm}; multi-disciplinary multimodal reasoning with MMMU-Pro~\citep{yue2025mmmu}; and logical reasoning with LogicVista~\citep{xiao2024logicvista}. We use exact-match evaluation against ground-truth answers and exclude datasets or instances with free-form answers that require LLM-as-a-judge evaluation. For all benchmarks, we report average accuracy@8 with an inference temperature of $1.0$, using a single fixed evaluation pipeline for all models\rev{, where accuracy@8 averages the pass@1 accuracy over eight independently sampled responses per question}.

\subsection{Benchmark Comparisons}
\label{main_exp}

Table~\ref{tab:main_reasoning_results} presents the main comparison with open-source reasoning LMMs. 
Under the same Qwen2.5-VL backbones, \rev{\EVAN{}} improves over the DAPO baseline from 50.3 to 51.6 average accuracy at the 3B scale and from 56.5 to 57.9 at the 7B scale. 
\nopagebreak This controlled comparison isolates the effect of evidence-ablated negatives, since \rev{\EVAN{}} keeps the clean-image RLVR pipeline while adding only a bounded auxiliary signal from evidence-ablated rollouts. 
The gains are also stable across scales: \rev{\EVAN{}} improves DAPO on eight of nine benchmarks for 3B and on most benchmarks for 7B. 
\rev{On Qwen3-VL-8B, \EVAN{} also improves its matched DAPO baseline from 71.9 to 73.0 average accuracy and outperforms it on eight of nine benchmarks, indicating that the gains extend to another model family.}
Compared with prior open-source reasoning LMMs, \rev{\EVAN{}} achieves the best average performance in \rev{each} size group, including a 0.9 point improvement over the strongest 7B baseline, VPPO-7B\rev{, and a 1.8 point improvement over VPPO-8B}.
These results support our main claim that plausible negatives induced by removing query-critical visual evidence provide useful training signal beyond standard outcome supervision, helping the policy reduce language-prior shortcuts while preserving strong clean-image reasoning performance.

\begin{table*}[t!]
\centering
\caption{Comparison with state-of-the-art open-source reasoning LMMs on multimodal reasoning benchmarks. All methods are evaluated under the same evaluation configuration. The best performance is marked in \textbf{bold}.}
\setlength{\tabcolsep}{3.5pt}
\renewcommand{\arraystretch}{1.2}
\resizebox{\textwidth}{!}{%
\begin{tabular}{lcccccccccc}
    \toprule
    \textbf{Model}
      & {\scriptsize\textbf{Geometry3K}}
      & {\scriptsize\textbf{MathVista}}
      & {\scriptsize\textbf{MathVerse}}
      & {\scriptsize\textbf{We-Math}}
      & {\scriptsize\textbf{DynaMath}}
      & {\scriptsize\textbf{MathVision}}
      & {\scriptsize\textbf{MMK12}}
      & {\scriptsize\textbf{MMMU-Pro}}
      & {\scriptsize\textbf{LogicVista}}
      & {\scriptsize\textbf{Avg.}}
      \\
    \midrule[0.6pt]
    \multicolumn{11}{c}{\rev{\textbf{Qwen2.5-VL-3B-Instruct}}} \\
    \midrule[0.6pt]
    PAPO-3B
      & 36.9 & \textbf{63.3} & 60.5 & 62.4 & 57.2 & 26.2 & 62.9 & 28.4 & 43.2 & 49.0 \\
    \rowcolor{gray!15}
    Qwen2.5-VL-3B
      & 19.4 & 46.3 & 33.5 & 29.9 & 40.4 & 18.2 & 36.3 & 19.8 & 32.1 & 30.7 \\
    {}+DAPO
      & 38.8 & 62.4 & 60.4 & 60.5 & 59.3 & 26.5 & \textbf{72.2} & 29.9 & 43.0 & 50.3 \\
    \rowcolor{blue!5}
    {}\textbf{+\rev{\EVAN{}} (ours)}
      & \textbf{39.9} & 62.7 & \textbf{62.8} & \textbf{64.6} & \textbf{60.6} & \textbf{27.6} & 71.5 & \textbf{31.0} & \textbf{43.3} & \textbf{51.6} \\
    \midrule[0.6pt]
    \multicolumn{11}{c}{\rev{\textbf{Qwen2.5-VL-7B-Instruct}}} \\
    \midrule[0.6pt]
    VL-Rethinker-7B
      & 39.5 & 61.0 & 67.7 & 67.7 & 61.1 & 30.3 & 64.5 & 36.5 & 45.5 & 52.6 \\
    R1-ShareVL-7B
      & 42.8 & 68.8 & 67.2 & 69.3 & 62.6 & 29.0 & 66.1 & 35.7 & 47.0 & 54.3 \\
    NoisyRollout-7B
      & 44.9 & 68.7 & 67.5 & \textbf{70.8} & 63.9 & 29.0 & 59.8 & 36.4 & 48.1 & 54.3 \\
    PAPO-7B
      & 44.5 & 67.9 & 68.5 & 68.2 & 64.8 & 28.8 & 79.6 & 36.2 & 47.5 & 56.2 \\
    \rev{BiPS-7B}
      & \rev{41.2} & \rev{\textbf{70.9}} & \rev{69.0} & \rev{70.2} & \rev{66.4} & \rev{30.7} & \rev{76.2} & \rev{37.7} & \rev{47.3} & \rev{56.6} \\
    VPPO-7B
      & 45.2 & 62.9 & \textbf{70.3} & 70.1 & 66.3 & 31.0 & 80.8 & 37.6 & 48.4 & 57.0 \\
    \rowcolor{gray!15}
    Qwen2.5-VL-7B
      & 36.1 & 63.9 & 38.6 & 46.6 & 56.1 & 18.6 & 43.4 & 25.9 & 42.8 & 41.3 \\
    {}+DAPO
      & 45.5 & 66.2 & 69.2 & 68.4 & 65.8 & 29.0 & 81.9 & 37.0 & 45.8 & 56.5 \\
    \rowcolor{blue!5}
    {}\textbf{+\rev{\EVAN{}} (ours)}
      & \textbf{46.0} & \rev{69.0} & 69.1 & 69.7 & \textbf{66.5} & \textbf{31.3} & \textbf{82.2} & \textbf{38.8} & \textbf{48.8} & \textbf{57.9} \\
    \midrule[0.6pt]
    \multicolumn{11}{c}{\rev{\textbf{Qwen3-VL-8B-Instruct}}} \\
    \midrule[0.6pt]
    \rev{VPPO-8B}
      & \rev{71.7} & \rev{70.3} & \rev{83.7} & \rev{83.6} & \rev{\textbf{78.9}} & \rev{56.3} & \rev{79.1} & \rev{\textbf{52.9}} & \rev{\textbf{64.0}} & \rev{71.2} \\
    \rowcolor{gray!15}
    \rev{Qwen3-VL-8B}
      & \rev{65.2} & \rev{68.7} & \rev{69.4} & \rev{77.2} & \rev{73.4} & \rev{38.0} & \rev{69.1} & \rev{40.3} & \rev{61.3} & \rev{62.5} \\
    \rev{{}+DAPO}
      & \rev{74.6} & \rev{71.9} & \rev{84.0} & \rev{83.9} & \rev{76.9} & \rev{56.7} & \rev{84.1} & \rev{52.3} & \rev{62.4} & \rev{71.9} \\
    \rowcolor{blue!5}
    \rev{{}\textbf{+\EVAN{} (ours)}}
      & \rev{\textbf{76.6}} & \rev{\textbf{74.9}} & \rev{\textbf{85.0}} & \rev{\textbf{84.9}} & \rev{77.6} & \rev{\textbf{57.0}} & \rev{\textbf{85.0}} & \rev{52.0} & \rev{63.6} & \rev{\textbf{73.0}} \\
    \bottomrule
\end{tabular}%
}
\vspace{-10pt}
\label{tab:main_reasoning_results}
\end{table*}

\subsection{Ablation Study}
\label{sec:ablation}

We conduct ablation studies on Qwen2.5-VL-3B to isolate the main components of \EVAN{}.
All variants use the same training and evaluation configuration.

\paragraph{\rev{RankMix} Selection.}
We ablate the selection strategy used by \EVAN{} under a fixed 20\% data budget, keeping the evidence-ablated image construction and policy optimization unchanged.
\rev{The Random variant uniformly samples 20\% of the training examples and then applies the same $\rho=50\%$ scaled-IIG-guided masking as the other variants.}
As shown in Table~\ref{tab:ablation_data_selection}, selecting examples only by global scaled-IIG underperforms random selection, suggesting that high visual dependency alone can reduce data diversity.
Category-wise selection improves the average score by preserving category coverage.
RankMix further improves the average score to 51.6 by balancing global visual dependency with category-level coverage.
This supports our choice of mixed ranking for selecting examples that are both visually dependent and sufficiently diverse.

\begin{table*}[t!]
\centering
\caption{Ablation study of \rev{RankMix} Selection with Qwen2.5-VL-3B. All variants use the same 20\% data budget and differ only in the selection strategy.}
\setlength{\tabcolsep}{3.5pt}
\renewcommand{\arraystretch}{1.2}
\resizebox{\textwidth}{!}{%
\begin{tabular}{lcccccccccc}
    \toprule
    \textbf{Variant}
      & {\scriptsize\textbf{Geometry3K}}
      & {\scriptsize\textbf{MathVista}}
      & {\scriptsize\textbf{MathVerse}}
      & {\scriptsize\textbf{We-Math}}
      & {\scriptsize\textbf{DynaMath}}
      & {\scriptsize\textbf{MathVision}}
      & {\scriptsize\textbf{MMK12}}
      & {\scriptsize\textbf{MMMU-Pro}}
      & {\scriptsize\textbf{LogicVista}}
      & {\scriptsize\textbf{Avg.}}
      \\
    \midrule[0.6pt]
    Random
      & 38.5 & 62.4 & 62.5 & \textbf{65.1} & 60.1 & 27.1 & 72.2 & 30.0 & 42.0 & 51.1 \\
    Global
      & 39.8 & 60.8 & 60.9 & 61.9 & 59.2 & 27.5 & 72.3 & 28.8 & 41.7 & 50.3 \\
    Category
      & 38.4 & \textbf{63.5} & \textbf{63.5} & 64.8 & 59.6 & 26.8 & \textbf{72.7} & 29.3 & 42.9 & 51.3 \\
    \rowcolor{blue!5}
    \textbf{RankMix (\rev{\EVAN{}})}
      & \textbf{39.9} & 62.7 & 62.8 & 64.6 & \textbf{60.6} & \textbf{27.6} & 71.5 & \textbf{31.0} & \textbf{43.3} & \textbf{51.6} \\
    \bottomrule
\end{tabular}%
}
\vspace{-10pt}
\label{tab:ablation_data_selection}
\end{table*}

\paragraph{Adversarial masking.}
We ablate the image construction strategy used for evidence-ablated rollouts. As shown in Table~\ref{tab:ablation_adversarial_image}, substituting the evidence-ablated image with the clean image in the auxiliary branch yields an average score of 50.7. This shows that extra rollouts alone explain only a small fraction of the gain of \EVAN{}. We further evaluate pixel-level perturbations by applying Fast Gradient Sign Method (FGSM)~\citep{goodfellow2014explaining,liu2025modality}, which only marginally improves the average score to 50.8. This suggests that pixel-level adversarial perturbations are insufficiently stable in the online RL setting, where the continuously evolving policy renders locally optimized perturbations rapidly uninformative. In contrast, the scaled-IIG-guided masking achieves an average score of 51.6, validating our design choice of employing patch-level evidence ablation to construct evidence-ablated images for hard negative generation. \rev{Replacing scaled-IIG-guided masking with random patch masking at the same $50\%$ ratio yields an average score of only 50.8, suggesting that the gain comes from localizing query-critical evidence rather than from the masked area alone.}

\begin{table*}[t!]
\centering
\caption{Ablation on evidence-ablated image construction with Qwen2.5-VL-3B. All variants use the same selected examples and differ only in the image used for evidence-ablated rollouts.}
\setlength{\tabcolsep}{3.5pt}
\renewcommand{\arraystretch}{1.2}
\resizebox{\textwidth}{!}{%
\begin{tabular}{lcccccccccc}
    \toprule
    \textbf{Variant}
      & {\scriptsize\textbf{Geometry3K}}
      & {\scriptsize\textbf{MathVista}}
      & {\scriptsize\textbf{MathVerse}}
      & {\scriptsize\textbf{We-Math}}
      & {\scriptsize\textbf{DynaMath}}
      & {\scriptsize\textbf{MathVision}}
      & {\scriptsize\textbf{MMK12}}
      & {\scriptsize\textbf{MMMU-Pro}}
      & {\scriptsize\textbf{LogicVista}}
      & {\scriptsize\textbf{Avg.}}
      \\
    \midrule[0.6pt]
    Clean Image
      & 39.8 & \rev{\textbf{63.5}} & 60.8 & 61.3 & 59.8 & 26.9 & 72.8 & 29.2 & 42.4 & 50.7 \\
    \rev{Random Patch}
      & \rev{\textbf{39.9}} & \rev{62.0} & \rev{62.1} & \rev{61.4} & \rev{59.6} & \rev{27.2} & \rev{72.9} & \rev{29.9} & \rev{42.8} & \rev{50.8} \\
    FGSM
      & 39.6 & 61.8 & 60.6 & 64.1 & 59.8 & 26.5 & \textbf{73.2} & 29.6 & 42.3 & 50.8 \\
    \rowcolor{blue!5}
    \textbf{Adv. mask (\EVAN{})}
      & \textbf{39.9} & 62.7 & \rev{\textbf{62.8}} & \textbf{64.6} & \textbf{60.6} & \textbf{27.6} & 71.5 & \textbf{31.0} & \textbf{43.3} & \textbf{51.6} \\
    \bottomrule
\end{tabular}%
}
\vspace{-10pt}
\label{tab:ablation_adversarial_image}
\end{table*}

\paragraph{\rev{Masking ratio.}}
\rev{We further vary the masking ratio $\rho$ of scaled-IIG-guided masking while keeping the training and evaluation settings unchanged.
As shown in Table~\ref{tab:ablation_mask_ratio}, masking $50\%$ of the patches achieves the best average score of 51.6, outperforming the $25\%$, $75\%$, and $100\%$ settings by 0.5, 0.2, and 0.7 points, respectively.
A lower ratio may leave part of the query-critical evidence visible, whereas $\rho=100\%$ blacks out the entire image while retaining the image tokens, so the evidence-ablated rollouts are generated without any visual evidence.
This fully black setting yields the lowest average score, consistent with our motivation in Section~\ref{sec:adv_mining} that overly destructive interventions may make the image uninformative and produce off-distribution failures.
These results support our choice of $\rho=50\%$, which removes query-critical evidence while preserving the remaining visual context.}

\begin{table*}[t!]
\centering
\color{revcolor}
\caption{Effect of the masking ratio $\rho$ on Qwen2.5-VL-3B. All variants use the same training recipe and differ only in the fraction of masked patches. $\rho=100\%$ masks the entire image.}
\setlength{\tabcolsep}{3.5pt}
\renewcommand{\arraystretch}{1.2}
\resizebox{\textwidth}{!}{%
\begin{tabular}{lcccccccccc}
    \toprule
    \textbf{Variant}
      & {\scriptsize\textbf{Geometry3K}}
      & {\scriptsize\textbf{MathVista}}
      & {\scriptsize\textbf{MathVerse}}
      & {\scriptsize\textbf{We-Math}}
      & {\scriptsize\textbf{DynaMath}}
      & {\scriptsize\textbf{MathVision}}
      & {\scriptsize\textbf{MMK12}}
      & {\scriptsize\textbf{MMMU-Pro}}
      & {\scriptsize\textbf{LogicVista}}
      & {\scriptsize\textbf{Avg.}}
      \\
    \midrule[0.6pt]
    \EVAN{} ($\rho=25\%$)
      & 39.6 & 59.9 & 61.5 & 64.9 & 61.0 & 26.8 & 72.7 & 31.1 & 42.8 & 51.1 \\
    \rowcolor{blue!5}
    \textbf{\EVAN{} ($\rho=50\%$)}
      & 39.9 & 62.7 & 62.8 & 64.6 & 60.6 & 27.6 & 71.5 & 31.0 & 43.3 & 51.6 \\
    \EVAN{} ($\rho=75\%$)
      & 39.4 & 61.6 & 62.2 & 63.7 & 60.2 & 27.9 & 73.0 & 30.7 & 43.5 & 51.4 \\
    \EVAN{} ($\rho=100\%$)
      & 39.5 & 63.2 & 63.1 & 65.0 & 59.1 & 26.5 & 71.3 & 29.4 & 41.0 & 50.9 \\
    \bottomrule
\end{tabular}%
}
\vspace{-10pt}
\label{tab:ablation_mask_ratio}
\end{table*}

\begin{figure*}[t!]
  \centering
  \includegraphics[width=1\linewidth]{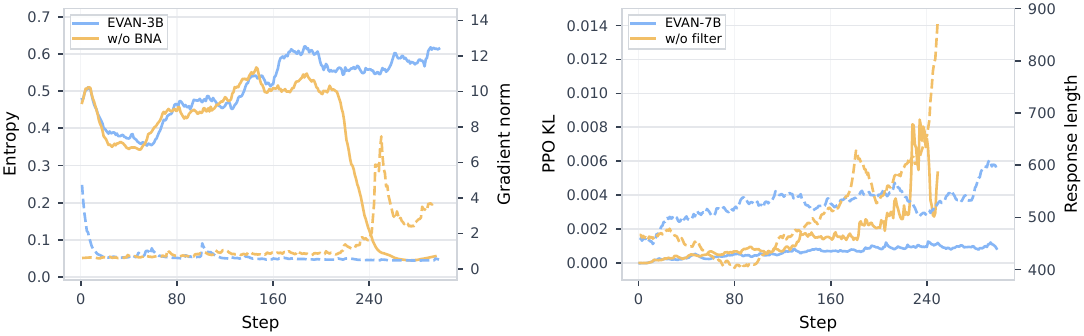}
  \caption{
(Left) Bounded negative advantage (BNA) ablation on Qwen2.5-VL-3B, with entropy on the left axis and gradient norm on the right axis.
(Right) rollout filtering ablation on Qwen2.5-VL-7B, with PPO KL Divergence on the left axis and response length on the right axis.
  }
  \label{fig:training_diagnostics}
\end{figure*}

\paragraph{Stability controls.}
We ablate the two stability controls used for evidence-ablated negatives.
As shown in Figure~\ref{fig:training_diagnostics}, removing Bounded negative advantage (BNA) causes late-stage entropy collapse and a sharp increase in gradient norm, indicating that the negative signal can dominate the clean RL update.
Removing rollout filtering also destabilizes training, with PPO KL and response length growing rapidly.
In contrast, \EVAN{} keeps these dynamics within stable ranges.
These results show that evidence-ablated negatives require both bounded advantages and clean-image plausibility filtering to provide useful training signal without destabilizing online RL.

\subsection{Further Analysis}
\label{sec:further_analysis}

We provide further analysis to better understand the behavior of \rev{\EVAN{}} during training.
Additional results on filtering behavior, selection ratio, and selected-data distribution are provided in the appendix.

\begin{figure*}[t!]
  \centering
  \includegraphics[width=1\linewidth]{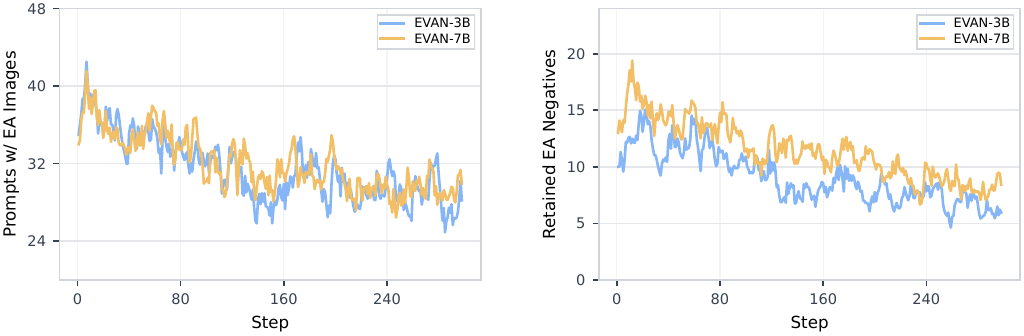}
  \caption{
  Training dynamics of the \rev{EVidence-Ablated Negatives} branch.
(Left) Number of prompts with prepared evidence-ablated images in each training step.
(Right) Number of retained evidence-ablated negative responses appended to the clean-image update after filtering.
  }
  \label{fig:ean-branch}
\end{figure*}

\paragraph{Evidence-ablated negatives.}
Figure~\ref{fig:ean-branch} shows the training dynamics of the evidence-ablated negatives.
Although evidence-ablated images are prepared for the selected 20\% of training examples, only a small number of evidence-ablated responses are retained after filtering and appended to the clean-image update.
This shows that \rev{\EVAN{}} uses a sparse auxiliary signal rather than adding many extra training responses.
Together with the improved benchmark performance, this suggests that the gain does not come from substantially increasing the amount of training data, but from using a small set of high-quality evidence-ablated negatives as additional supervision.

%% file: sec/appendix.tex
\onecolumn
\begin{center}
    {\LARGE \bf Appendix}
\end{center}

\section{Experimental Details}
\subsection{Training Details}
We use DAPO~\citep{yu2025dapo} as both the baseline method and the underlying optimizer for \EVAN{}. 
For all experiments, we disable the KL penalty by setting the KL coefficient to $0$, and use asymmetric clipping with $\epsilon_{\mathrm{low}}=0.2$ and $\epsilon_{\mathrm{high}}=0.28$. 
The maximum response length is set to $4{,}096$ tokens \rev{for the 3B and 7B models and $8{,}192$ tokens for the 8B model}.
At each rollout step, we generate responses for $256$ prompts and optimize the policy with a mini-batch size of $32$, which gives $8$ gradient updates per rollout step. 
Training runs for $300$ prompt-generation steps, corresponding to $2$ epochs over the training dataset. 
For \EVAN{}, we generate $2$ responses for each evidence-ablated image \rev{with the 3B model and $4$ responses with the 7B and 8B models} to obtain auxiliary  candidates. Experiments are conducted on 4 $\times$ NVIDIA H200 GPUs, and \rev{the} training run of Qwen2.5-VL-7B takes \rev{36.05 hours with DAPO and 36.74 hours with \EVAN{}}.

\subsection{Training Prompt and Reward}
\label{app:training_prompt_reward}

For all training rollouts, we append a fixed instruction to each question to standardize the response format:
\begin{quote}
\small
\texttt{You first think through the reasoning process as an internal monologue, enclosed within <think> </think> tags. Then, provide your final answer enclosed within \textbackslash boxed\{\}.}
\end{quote}
This instruction encourages the model to separate its reasoning process from the final answer, and also makes answer extraction more reliable.

The reward contains an accuracy term and a format term. 
The accuracy term checks whether the final boxed answer matches the ground-truth answer, while the format term checks whether the response follows the required structure with \texttt{<think>} and \texttt{</think>} tags and a final \verb|\boxed{}| answer. 
The total reward is computed as
\begin{equation}
R(o) = 0.9 \cdot R_{\mathrm{acc}}(o) + 0.1 \cdot R_{\mathrm{fmt}}(o),
\end{equation}
where $R_{\mathrm{acc}}(o)$ denotes the accuracy score and $R_{\mathrm{fmt}}(o)$ denotes the format score. 
This reward design prioritizes answer correctness while keeping a small incentive for producing parseable responses.

\subsection{\texorpdfstring{\rev{Bounded Negative Advantage under Binary Rewards}}{Bounded Negative Advantage under Binary Rewards}}
\label{app:bounded_adv}
\rev{Consider binary rewards and let $p$ be the fraction of correct responses in the clean rollout group, where dynamic sampling keeps $0<p<1$.
The group mean is $p$ and the group standard deviation is $\sqrt{p(1-p)}$, so Eq.~\eqref{eq:advantage} assigns $\sqrt{(1-p)/p}$ to each correct response and $-\sqrt{p/(1-p)}$ to each incorrect response.
Eq.~\eqref{eq:adv_advantage} therefore gives
\begin{equation}
\label{eq:bounded_adv_binary}
\hat A_j^{a}=-\min\left\{\sqrt{\frac{1-p}{p}},\ \sqrt{\frac{p}{1-p}}\right\}.
\end{equation}
When most clean rollouts are incorrect ($p<0.5$), the evidence-ablated negative receives the same advantage as a clean incorrect response.
When most clean rollouts are correct ($p>0.5$), its magnitude equals the advantage of a clean correct response and is smaller than the penalty on a clean incorrect response.
The magnitude reaches its maximum of $1$ at $p=0.5$, where the clean outcomes are most uncertain, and approaches $0$ as $p$ approaches $0$ or $1$, where the behavior of the model on the example is already stable.
The same values hold for the reward in Appendix~\ref{app:training_prompt_reward} whenever all responses in the group receive the same format score, because the rewards then take only two values and the advantage in Eq.~\eqref{eq:advantage} does not change when a constant is added to all rewards or when all rewards are multiplied by a positive constant.
Table~\ref{tab:bounded_adv} lists these values for the group size $G=8$ used in our experiments.}

\begin{table*}[h]
\centering
\color{revcolor}
\caption{Advantages under binary rewards for a clean rollout group of $G=8$ responses. Each column corresponds to a number of correct responses in the clean rollout group, and each row gives the advantage assigned to one type of response.}
\label{tab:bounded_adv}
\setlength{\tabcolsep}{6pt}
\renewcommand{\arraystretch}{1.2}
\begin{tabular}{lccccccc}
    \toprule
    \textbf{Correct clean responses}
      & \textbf{1} & \textbf{2} & \textbf{3} & \textbf{4} & \textbf{5} & \textbf{6} & \textbf{7}
      \\
    \midrule[0.6pt]
    Clean correct response
      & 2.65 & 1.73 & 1.29 & 1.00 & 0.77 & 0.58 & 0.38 \\
    Clean incorrect response
      & $-0.38$ & $-0.58$ & $-0.77$ & $-1.00$ & $-1.29$ & $-1.73$ & $-2.65$ \\
    Evidence-ablated negative
      & $-0.38$ & $-0.58$ & $-0.77$ & $-1.00$ & $-0.77$ & $-0.58$ & $-0.38$ \\
    \bottomrule
\end{tabular}
\end{table*}

\section{Additional Experiments}
\subsection{\texorpdfstring{\rev{Confidence intervals of main results.}}{Confidence intervals of main results.}}
\label{app:confidence_intervals}
\rev{To assess the stability of the main results in Table~\ref{tab:main_reasoning_results}, Table~\ref{tab:confidence_intervals} reports 95\% confidence intervals for DAPO and \EVAN{} on every benchmark at both the 3B and 7B scales.
Each interval is a Student's $t$ interval with seven degrees of freedom, computed from the eight stochastic pass@1 evaluations underlying accuracy@8 for the same fixed checkpoint.
The interval of the average score is computed in the same way from the eight per-evaluation averages over all nine benchmarks.
\EVAN{} outperforms DAPO in average accuracy at both scales, and its 95\% confidence interval lies entirely above that of DAPO, with $51.6\pm0.18$ versus $50.3\pm0.25$ for 3B and $57.9\pm0.23$ versus $56.5\pm0.16$ for 7B.
This shows that the overall gains are stable under stochastic decoding.}

\begin{table*}[h]
\centering
\color{revcolor}
\caption{95\% confidence intervals of DAPO and \EVAN{} on Qwen2.5-VL-3B and Qwen2.5-VL-7B. Each interval is computed from the eight stochastic pass@1 evaluations of the same checkpoint.}
\setlength{\tabcolsep}{2pt}
\renewcommand{\arraystretch}{1.2}
\resizebox{\textwidth}{!}{%
\begin{tabular}{lcccccccccc}
    \toprule
    \textbf{Model}
      & {\scriptsize\textbf{Geometry3K}}
      & {\scriptsize\textbf{MathVista}}
      & {\scriptsize\textbf{MathVerse}}
      & {\scriptsize\textbf{We-Math}}
      & {\scriptsize\textbf{DynaMath}}
      & {\scriptsize\textbf{MathVision}}
      & {\scriptsize\textbf{MMK12}}
      & {\scriptsize\textbf{MMMU-Pro}}
      & {\scriptsize\textbf{LogicVista}}
      & {\scriptsize\textbf{Avg.}}
      \\
    \midrule[0.6pt]
    \multicolumn{11}{c}{\textbf{Qwen2.5-VL-3B-Instruct}} \\
    \midrule[0.6pt]
    DAPO
      & 38.8{\scriptsize$\pm$0.80} & 62.4{\scriptsize$\pm$0.52} & 60.4{\scriptsize$\pm$0.28} & 60.5{\scriptsize$\pm$0.47} & 59.3{\scriptsize$\pm$0.29} & 26.5{\scriptsize$\pm$0.26} & 72.2{\scriptsize$\pm$0.24} & 29.9{\scriptsize$\pm$0.46} & 43.0{\scriptsize$\pm$0.93} & 50.3{\scriptsize$\pm$0.25} \\
    \rowcolor{blue!5}
    \textbf{\EVAN{}}
      & 39.9{\scriptsize$\pm$0.66} & 62.7{\scriptsize$\pm$0.45} & 62.8{\scriptsize$\pm$0.50} & 64.6{\scriptsize$\pm$0.42} & 60.6{\scriptsize$\pm$0.17} & 27.6{\scriptsize$\pm$0.58} & 71.5{\scriptsize$\pm$0.40} & 31.0{\scriptsize$\pm$0.61} & 43.3{\scriptsize$\pm$1.14} & 51.6{\scriptsize$\pm$0.18} \\
    \midrule[0.6pt]
    \multicolumn{11}{c}{\textbf{Qwen2.5-VL-7B-Instruct}} \\
    \midrule[0.6pt]
    DAPO
      & 45.5{\scriptsize$\pm$0.99} & 66.2{\scriptsize$\pm$0.58} & 69.2{\scriptsize$\pm$0.46} & 68.4{\scriptsize$\pm$0.32} & 65.8{\scriptsize$\pm$0.12} & 29.0{\scriptsize$\pm$0.26} & 81.9{\scriptsize$\pm$0.24} & 37.0{\scriptsize$\pm$0.44} & 45.8{\scriptsize$\pm$1.19} & 56.5{\scriptsize$\pm$0.16} \\
    \rowcolor{blue!5}
    \textbf{\EVAN{}}
      & 46.0{\scriptsize$\pm$1.08} & 69.0{\scriptsize$\pm$0.46} & 69.1{\scriptsize$\pm$0.58} & 69.7{\scriptsize$\pm$0.33} & 66.5{\scriptsize$\pm$0.45} & 31.3{\scriptsize$\pm$0.35} & 82.2{\scriptsize$\pm$0.24} & 38.8{\scriptsize$\pm$0.50} & 48.8{\scriptsize$\pm$1.07} & 57.9{\scriptsize$\pm$0.23} \\
    \bottomrule
\end{tabular}%
}
\vspace{-10pt}
\label{tab:confidence_intervals}
\end{table*}

\subsection{\texorpdfstring{\rev{Transfer beyond structured reasoning.}}{Transfer beyond structured reasoning.}}
\label{app:transfer}
\rev{To evaluate whether the gains of \EVAN{} extend beyond structured reasoning, Table~\ref{tab:transfer} reports the results of the Qwen2.5-VL-3B DAPO and \EVAN{} checkpoints on the full ChartQA~\citep{masry2022chartqa}, GQA~\citep{hudson2019gqa}, and Visual CounterFact~\citep{golovanevsky2025pixels} evaluation sets.
Each score is the mean sampled pass@1 over eight evaluations, using relaxed accuracy for ChartQA and exact-match accuracy for GQA.
For Visual CounterFact, we use the direct counterfactual-grounding condition (CF+this), in which a counterfactual image is paired with an instance-specific question and the correct answer follows the modified visual attribute rather than the corresponding world-knowledge prior.
\EVAN{} outperforms DAPO on all three benchmarks and improves the pooled Visual CounterFact accuracy from 81.61 to 85.01.
These results show that \EVAN{} transfers beyond structured reasoning to chart understanding, general visual question answering, and counterfactual visual grounding.}

\begin{table*}[h]
\centering
\color{revcolor}
\caption{Transfer of the Qwen2.5-VL-3B DAPO and \EVAN{} checkpoints beyond structured reasoning.}
\setlength{\tabcolsep}{8pt}
\renewcommand{\arraystretch}{1.2}
\begin{tabular}{lcc*{3}{>{\centering\arraybackslash}p{1.5cm}}}
    \toprule
    \textbf{Model}
      & \textbf{ChartQA}
      & \textbf{GQA}
      & \multicolumn{3}{c}{\textbf{Visual CounterFact (CF+this)}}
      \\
    \cmidrule(lr){4-6}
      &
      &
      & Color
      & Size
      & Pooled
      \\
    \midrule[0.6pt]
    DAPO
      & 81.74 & 54.46 & 84.61 & 79.57 & 81.61 \\
    \rowcolor{blue!5}
    \textbf{\EVAN{}}
      & 82.55 & 55.83 & 85.04 & 84.99 & 85.01 \\
    \bottomrule
\end{tabular}
\vspace{-8pt}
\label{tab:transfer}
\end{table*}

\subsection{\texorpdfstring{\rev{Computational cost.}}{Computational cost.}}
\label{app:compute_cost}
\rev{Table~\ref{tab:compute_cost} compares the wall-clock cost of the matched 300-step Qwen2.5-VL-7B runs of DAPO and \EVAN{} on the same 4 $\times$ NVIDIA H200 GPUs.
The rollout time of \EVAN{} includes both the standard clean-image generation and the auxiliary evidence-ablated generation.
Filtering the evidence-ablated candidates is a lightweight host-side selection that takes negligible time relative to generation and is included in the total training time.
\EVAN{} increases the total training time by 1.9\%, and its cumulative rollout time is lower than that of DAPO because the sampled response lengths and the number of dynamic-sampling batches differ between the two runs.
In addition, \EVAN{} requires a one-time offline preprocessing step, which takes about 2.89 hours on 8 $\times$ NVIDIA H200 GPUs for scaled-IIG scoring and 1.98 hours on 4 $\times$ NVIDIA H200 GPUs for scaled-IIG-guided mask construction.
The auxiliary branch is removed after training, so \EVAN{} adds no parameters or computation at inference time.}

\begin{table*}[h]
\centering
\color{revcolor}
\caption{Training cost of the matched 300-step Qwen2.5-VL-7B runs on 4 $\times$ NVIDIA H200 GPUs, excluding the one-time preprocessing of \EVAN{}.}
\setlength{\tabcolsep}{8pt}
\renewcommand{\arraystretch}{1.2}
\begin{tabular}{lcc}
    \toprule
    \textbf{Method}
      & \textbf{Total training time (h)}
      & \textbf{Cumulative rollout time (h)}
      \\
    \midrule[0.6pt]
    DAPO
      & 36.05 & $\approx$22.0 \\
    \rowcolor{blue!5}
    \textbf{\EVAN{}}
      & 36.74 & $\approx$21.2 \\
    \bottomrule
\end{tabular}
\vspace{-8pt}
\label{tab:compute_cost}
\end{table*}

\subsection{Rollout filtering ratio during training.}
\begin{figure*}[h]
  \centering
  \includegraphics[width=1\linewidth]{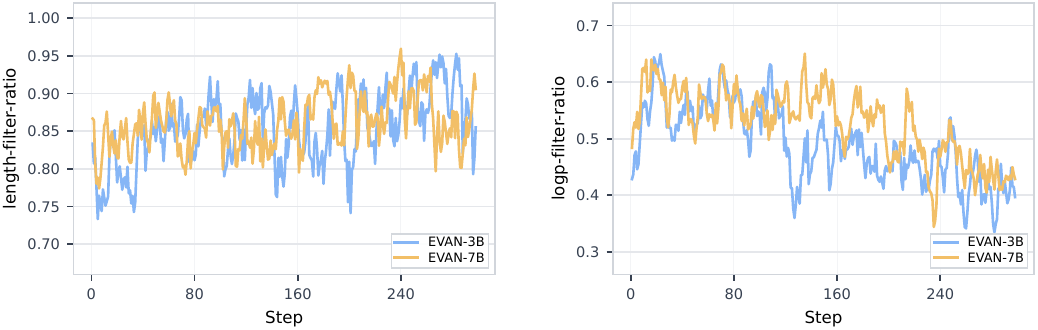}
  \caption{
  Rollout filtering ratios during training. The fraction of  candidates that pass the length filter (left) and the clean-image likelihood filter (right) for \rev{\EVAN{}}-3B and \rev{\EVAN{}}-7B.
  }
  \label{fig:filter}
\end{figure*}

As illustrated in Figure~\ref{fig:filter}, both the length filter and the log-probability (logp) filter play a crucial role in curating high-quality training signals. The length filter (left) consistently ensures that evidence-ablated negatives remain comparable in scale to clean rollouts, while the logp filter (right) effectively rejects evidence-ablated responses that are out-of-distribution or lack plausibility under the clean visual context. The stable filtering ratios observed across both 3B and 7B models demonstrate that our approach effectively suppresses low-quality or unstable gradients by filtering out uninformative failure cases.

\subsection{\texorpdfstring{\rev{Quality of retained negatives.}}{Quality of retained negatives.}}
\label{app:negative_quality}
\rev{Masking query-critical patches intentionally makes the original answer unverifiable, which is the condition under which a language-biased policy still answers from priors.
To verify that the retained negatives remain task-directed rather than describing the mask, we audit the evidence-ablated responses of the \EVAN{} run on Qwen2.5-VL-3B, where 3,906 of the 22,864 evidence-ablated candidates, or 17.1\%, pass all filters and are used for training.
We use GPT-5.5 to inspect 250 retained responses and 100 responses removed by the clean-image likelihood filter, and we also manually review 50 retained responses.
As shown in Table~\ref{tab:negative_quality}, all retained responses remain semantically relevant and task-directed in both audits, and none describes the mask or refuses to answer because the visual evidence is obscured.
In contrast, 40 of the 100 responses removed by the likelihood filter are pathological, with 30 containing degenerate text and 10 explicitly describing the mask.
These results show that the likelihood filter removes mask-induced pathological responses and keeps plausible task-directed failures as negatives.}

\begin{table*}[h]
\centering
\color{revcolor}
\caption{Semantic audit of evidence-ablated responses from the \EVAN{} run on Qwen2.5-VL-3B. Retained responses pass all filters and are used for training. Pathological responses contain degenerate text or explicitly describe the mask.}
\setlength{\tabcolsep}{6pt}
\renewcommand{\arraystretch}{1.2}
\resizebox{\textwidth}{!}{%
\begin{tabular}{llcccc}
    \toprule
    \textbf{Responses}
      & \textbf{Reviewer}
      & \textbf{Audited}
      & \textbf{Pathological}
      & \textbf{Degenerate text}
      & \textbf{Describes mask}
      \\
    \midrule[0.6pt]
    Retained
      & GPT-5.5 & 250 & 0 & 0 & 0 \\
    Retained
      & Human & 50 & 0 & 0 & 0 \\
    Removed by likelihood filter
      & GPT-5.5 & 100 & 40 & 30 & 10 \\
    \bottomrule
\end{tabular}%
}
\vspace{-8pt}
\label{tab:negative_quality}
\end{table*}

\subsection{Effect of selection ratio.}
We investigate the impact of the RankMix selection ratio $\tau$ on model performance. As shown in \rev{Table~\ref{tab:analysis_selection_ratio}}, \rev{\EVAN{}} achieves its best average performance with a sparse selection ratio of $\tau=20\%$. Notably, increasing the ratio to $30\%$ leads to a performance degradation, with the average score dropping to $50.8$. This trend suggests that applying evidence-ablated negatives to an excessively large portion of the dataset may be counterproductive. It indicates that the strategy of compressing language priors should be selectively applied only to examples with high visual dependency. These results further validate our \rev{scaled-IIG-guided} mining approach, demonstrating that focusing on a small but high-quality subset of visually dependent data is more effective than uniform intervention across the entire dataset.  

\begin{table*}[h]
\centering
\caption{Effect of the RankMix selection ratio $\tau$ on Qwen2.5-VL-3B. All variants use the same training recipe and differ only in the fraction of selected visually dependent examples.}
\setlength{\tabcolsep}{3.5pt}
\renewcommand{\arraystretch}{1.2}
\resizebox{\textwidth}{!}{%
\begin{tabular}{lcccccccccc}
    \toprule
    \textbf{Variant}
      & {\scriptsize\textbf{Geometry3K}}
      & {\scriptsize\textbf{MathVista}}
      & {\scriptsize\textbf{MathVerse}}
      & {\scriptsize\textbf{We-Math}}
      & {\scriptsize\textbf{DynaMath}}
      & {\scriptsize\textbf{MathVision}}
      & {\scriptsize\textbf{MMK12}}
      & {\scriptsize\textbf{MMMU-Pro}}
      & {\scriptsize\textbf{LogicVista}}
      & {\scriptsize\textbf{Avg.}}
      \\
    \midrule[0.6pt]
    \EVAN{} ($\tau=10\%$)
      & 39.5 & 63.2 & 63.2 & 63.9 & 59.3 & 27.4 & 72.4 & 30.1 & 44.6 & 51.5 \\
    \rowcolor{blue!5}
    \textbf{\EVAN{} ($\tau=20\%$)}
      & 39.9 & 62.7 & 62.8 & 64.6 & 60.6 & 27.6 & 71.5 & 31.0 & 43.3 & 51.6 \\
    \EVAN{} ($\tau=30\%$)
      & 38.8 & 63.7 & 60.3 & 60.3 & 61.1 & 27.5 & 72.1 & 30.3 & 43.0 & 50.8 \\
    \bottomrule
\end{tabular}%
}
\vspace{-10pt}
\label{tab:analysis_selection_ratio}
\end{table*}

\subsection{Scaled-IIG Change after Image Perturbation}

Table~\ref{tab:scaled_iig_attack_change} evaluates how image perturbation changes the visual-dependency signal on the RankMix-selected data.
Adversarial masking reduces both scaled-IIG and model accuracy after intervention, indicating that it weakens the visual evidence used by the model while retaining a non-trivial level of task solvability.
In contrast, FGSM causes a much sharper drop in both metrics, suggesting that pixel-level perturbation is overly destructive for online negative generation.
These results support the use of adversarial masking as a controlled intervention for constructing evidence-ablated images.

\begin{table*}[h]
\centering
\caption{Scaled-IIG and model accuracy before and after image perturbation on the RankMix 20\% selected data.}
\setlength{\tabcolsep}{8pt}
\renewcommand{\arraystretch}{1.12}
\resizebox{0.8\textwidth}{!}{%
\begin{tabular}{lcccc}
    \toprule
    \textbf{Variant}
      & \textbf{\rev{Scaled-IIG} Before}
      & \textbf{\rev{Scaled-IIG} After}
      & \textbf{Acc. Before (\%)}
      & \textbf{Acc. After (\%)}
      \\
    \midrule[0.6pt]
    Adv. mask
      & 0.595 & 0.455 & 88.3 & 66.3 \\
    FGSM
      & 0.595 & 0.093 & 88.3 & 35.3 \\
    \bottomrule
\end{tabular}%
}
\vspace{-8pt}
\label{tab:scaled_iig_attack_change}
\end{table*}

\subsection{\texorpdfstring{\rev{Scaled-IIG and response correctness.}}{Scaled-IIG and response correctness.}}
\label{app:scaled_iig_correctness}
\rev{We further examine whether scaled-IIG reflects response correctness on 5,000 ViRL39K questions, sampling five responses per question from each of Qwen2.5-VL-3B-Instruct, Qwen2.5-VL-7B-Instruct, Qwen3-VL-8B-Instruct, and Qwen3-VL-8B-Thinking.
As shown in Table~\ref{tab:scaled_iig_correctness}, using the scaled-IIG of each response to predict its correctness yields an AUC above 0.5 for every model, with a mean of 0.585.
Responses in the highest scaled-IIG quintile are also more accurate than those in the lowest quintile for every model, with a mean accuracy of 69.9\% versus 52.9\%, a 17.0 point gap.
These results show that scaled-IIG is positively associated with response correctness and support its use as a visual-dependency signal.}

\begin{table*}[h]
\centering
\color{revcolor}
\caption{Scaled-IIG and response correctness on 5,000 ViRL39K questions with five sampled responses per question. AUC measures how well the scaled-IIG of each response predicts its correctness. Top 20\% and Bottom 20\% report the accuracy (\%) of the responses with the highest and lowest scaled-IIG, and Gap is their difference.}
\setlength{\tabcolsep}{8pt}
\renewcommand{\arraystretch}{1.2}
\begin{tabular}{lcccc}
    \toprule
    \textbf{Model}
      & \textbf{AUC}
      & \textbf{Top 20\%}
      & \textbf{Bottom 20\%}
      & \textbf{Gap}
      \\
    \midrule[0.6pt]
    Qwen2.5-VL-3B-Instruct
      & 0.566 & 43.6 & 28.5 & 15.1 \\
    Qwen2.5-VL-7B-Instruct
      & 0.583 & 63.5 & 41.0 & 22.6 \\
    Qwen3-VL-8B-Instruct
      & 0.592 & 85.8 & 71.7 & 14.1 \\
    Qwen3-VL-8B-Thinking
      & 0.600 & 86.6 & 70.5 & 16.2 \\
    \midrule
    Mean
      & 0.585 & 69.9 & 52.9 & 17.0 \\
    \bottomrule
\end{tabular}
\vspace{-8pt}
\label{tab:scaled_iig_correctness}
\end{table*}

\subsection{Distribution of selected data.}
Table~\ref{tab:selected_data_distribution} compares the category distribution of the full training set, the globally ranked subset, and the subsets selected by RankMix.
Global top-20\% selection concentrates heavily on a few \rev{categories with high scaled-IIG}, especially Tables/Charts, while substantially reducing coverage of other categories.
In contrast, RankMix preserves a much broader category distribution across different selection ratios and remains closer to the full training pool.
This result supports our design choice that mixed global and category-wise ranking can focus on visually dependent examples without causing severe diversity collapse.

\begin{table*}[h]
\centering
\caption{Category distribution of selected training examples.}
\setlength{\tabcolsep}{3.5pt}
\renewcommand{\arraystretch}{1.15}
\resizebox{\textwidth}{!}{%
\begin{tabular}{lccccccccc}
    \toprule
    \textbf{Selection}
      & {\scriptsize\textbf{Non-Geo Math}}
      & {\scriptsize\textbf{Geometric}}
      & {\scriptsize\textbf{Tables/Charts}}
      & {\scriptsize\textbf{Science}}
      & {\scriptsize\textbf{Spatial}}
      & {\scriptsize\textbf{Broader STEM}}
      & {\scriptsize\textbf{Social Science}}
      & {\scriptsize\textbf{Commonsense}}
      & {\scriptsize\textbf{Total}}
      \\
    \midrule[0.6pt]
    Full dataset
      & 14901 & 10631 & 6189 & 3475 & 1922 & 956 & 676 & 120 & 38870 \\
    RankMix 10\%
      & 1122 & 382 & 1349 & 428 & 298 & 164 & 121 & 23 & 3887 \\
    RankMix 20\%
      & 2471 & 1131 & 2235 & 857 & 521 & 304 & 214 & 41 & 7774 \\
    RankMix 30\%
      & 3860 & 2092 & 2974 & 1241 & 720 & 423 & 295 & 56 & 11661 \\
    Global Top-20\%
      & 1244 & 306 & 4125 & 720 & 529 & 463 & 307 & 80 & 7774 \\
    \bottomrule
\end{tabular}%
}
\vspace{-10pt}
\label{tab:selected_data_distribution}
\end{table*}

\subsection{Case Study}
\label{app:case_study}

\begin{figure}[t]
    \centering
    \includegraphics[width=0.72\linewidth]{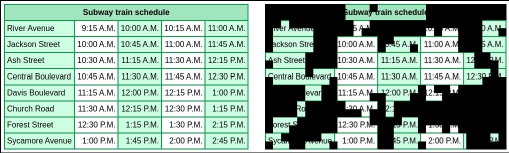}
    \caption{Case study on a train-schedule question. Left: clean image. Right: evidence-ablated image after adversarial masking.}
    \label{fig:case_study_schedule}
\end{figure}

\begin{figure}[h]
    \centering
    \color{revcolor}
    \includegraphics[width=0.7\linewidth]{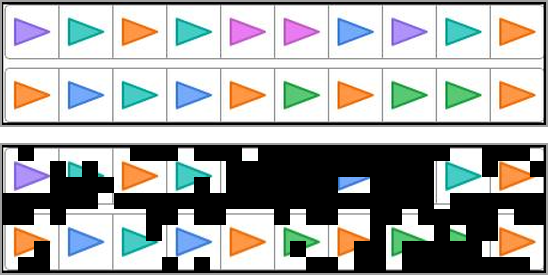}
    \caption{Case study on a counting question. Top: clean image. Bottom: evidence-ablated image after adversarial masking.}
    \label{fig:case_study_triangles}
\end{figure}

\begin{figure}[h]
    \centering
    \color{revcolor}
    \includegraphics[width=0.6\linewidth]{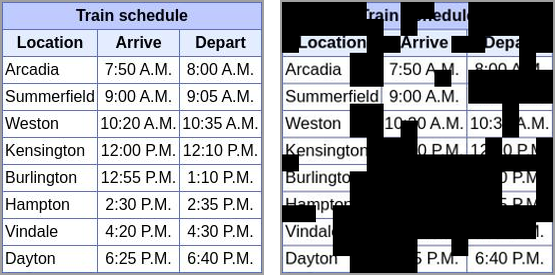}
    \caption{Case study on a table question. Left: clean image. Right: evidence-ablated image after adversarial masking.}
    \label{fig:case_study_burlington}
\end{figure}

\begin{figure}[h]
    \centering
    \color{revcolor}
    \includegraphics[width=0.9\linewidth]{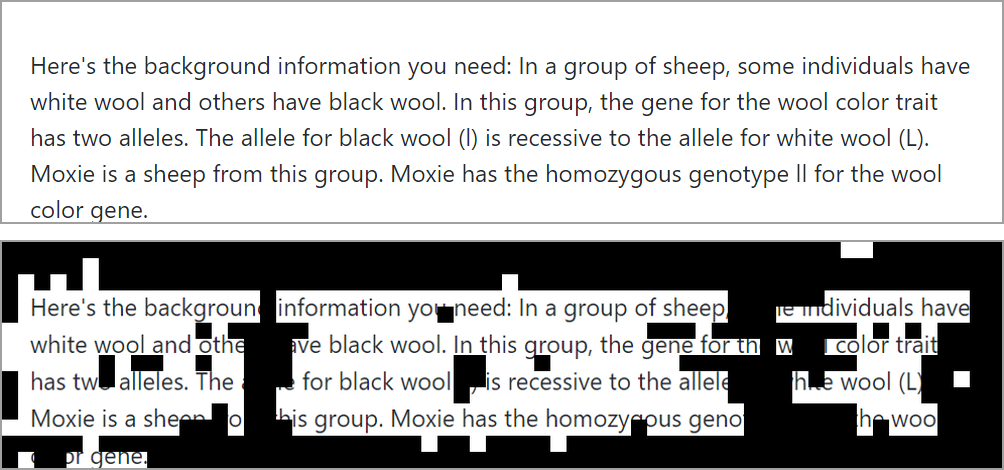}
    \caption{Case study on a genetics question. Top: clean image. Bottom: evidence-ablated image after adversarial masking.}
    \label{fig:case_study_genotype}
\end{figure}

\rev{Figures~\ref{fig:case_study_schedule}, \ref{fig:case_study_triangles}, \ref{fig:case_study_burlington}, and \ref{fig:case_study_genotype} show representative examples of train-schedule, counting, table, and genetics questions, respectively.
The images in Figures~\ref{fig:case_study_schedule}, \ref{fig:case_study_triangles}, and \ref{fig:case_study_burlington} are densely filled with content, and the image in Figure~\ref{fig:case_study_genotype} contains large blank areas.
For Figure~\ref{fig:case_study_schedule}, we show the responses of Qwen2.5-VL-3B and Qwen2.5-VL-7B on the evidence-ablated image.
For Figures~\ref{fig:case_study_triangles}, \ref{fig:case_study_burlington}, and \ref{fig:case_study_genotype}, the response is generated by the Qwen2.5-VL-3B \EVAN{} policy on the evidence-ablated image and passes both the likelihood filter and the length filter, so it is used as a negative in the clean-image update.}

\paragraph{Question \rev{of Figure~\ref{fig:case_study_schedule}}.}
\begin{quote}\small
Look at the train schedule. Colleen just missed the 10:15 AM train at River Avenue. Calculate the waiting time until the next train arrives.
\end{quote}

\paragraph{Qwen2.5-VL-3B on the evidence-ablated image.}
\begin{quote}\footnotesize\ttfamily
<think>\\
The table shows the train schedule at River Avenue.\\
Look at the first two trains on the left side of the table,\\
the first train departs at 10:00 AM and the second train departs at 10:15 AM.\\
So, the waiting time until the next train arrives is from 00:00 to 00:15.\\
Therefore, the waiting time is 15 minutes.\\
The waiting time is 15 minutes.\\
</think>\\
\textbackslash boxed\{15\}
\end{quote}

\paragraph{Qwen2.5-VL-7B on the evidence-ablated image.}
\begin{quote}\footnotesize\ttfamily
<think>\\
The train schedule shows the times for a freight train at different stations. Colleen missed the 10:15 AM train at River Avenue. To find the waiting time until the next train arrives, I need to look at the next station in the schedule, which is Jackson Street.\\
\\
The next train at Jackson Street is scheduled for 10:45 A.M., which is 30 minutes after the missed train at River Avenue.\\
\\
Therefore, the waiting time until the next train arrives is 30 minutes.\\
</think>\\
\\
\textbackslash boxed\{30 minutes\}
\end{quote}

\paragraph{\rev{Counting question of Figure~\ref{fig:case_study_triangles}.}}
\begin{quote}\small\color{revcolor}
How many triangles are there of each color, and what is the total number of triangles?
\end{quote}

\paragraph{\rev{Retained negative from Qwen2.5-VL-3B.}}
\begin{quote}\footnotesize\ttfamily\raggedright\color{revcolor}
<think>\\
To determine the number of triangles of each color and the total number of triangles in the given sequence, we need to carefully count the triangles of each color. Let's analyze the sequence step by step:\\[0.5\baselineskip]
1. **Identify the colors and count the triangles:**\\
\hspace*{1.5em}- Purple triangles: $\triangle$, $\triangle$\\
\hspace*{1.5em}- Orange triangles: $\triangle$, $\triangle$, $\triangle$\\
\hspace*{1.5em}- Teal triangles: $\triangle$, $\triangle$, $\triangle$, $\triangle$, $\triangle$, $\triangle$, $\triangle$\\
\hspace*{1.5em}- Blue triangles: $\triangle$, $\triangle$\\
\hspace*{1.5em}- Green triangles: $\triangle$\\[0.5\baselineskip]
2. **Count each color:**\\
\hspace*{1.5em}- Purple triangles: There are 2 purple triangles.\\
\hspace*{1.5em}- Orange triangles: There are 3 orange triangles.\\
\hspace*{1.5em}- Teal triangles: There are 7 teal triangles.\\
\hspace*{1.5em}- Blue triangles: There are 2 blue triangles.\\
\hspace*{1.5em}- Green triangles: There are 2 green triangles.\\[0.5\baselineskip]
3. **Summarize the count:**\\
\hspace*{1.5em}- Purple: 2\\
\hspace*{1.5em}- Orange: 3\\
\hspace*{1.5em}- Teal: 7\\
\hspace*{1.5em}- Blue: 2\\
\hspace*{1.5em}- Green: 2\\[0.5\baselineskip]
Now, we sum these counts to find the total number of triangles:
\[ 2 (purple) + 3 (orange) + 7 (teal) + 2 (blue) + 2 (green) = 16 \]
So the total number of triangles is 16.\\
</think>\\
\textbackslash{}boxed\{16\}
\end{quote}

\paragraph{\rev{Table question of Figure~\ref{fig:case_study_burlington}.}}
\begin{quote}\small\color{revcolor}
How long is the train scheduled to stop at Burlington?
\end{quote}

\paragraph{\rev{Retained negative from Qwen2.5-VL-3B.}}
\begin{quote}\footnotesize\ttfamily\raggedright\color{revcolor}
<think>\\
To determine the length of the train stop at Burlington, we need to find the difference between the departure time and the arrival time at Burlington. From the schedule:\\
- The train arrives at Burlington at 11:00 P.M.\\
- The train departs from Burlington at 10:00 P.M.\\[0.5\baselineskip]
The difference between these two times will give us the duration of the stop. We can calculate it as follows:\\
- Departure time: 10:00 P.M.\\
- Arrival time: 11:00 P.M.\\
- The difference between 11:00 P.M. and 10:00 P.M. is 1 hour.\\[0.5\baselineskip]
So, the train is scheduled to stop for 1 hour at Burlington.\\
</think>\\
\textbackslash{}boxed\{1 \textbackslash{}text\{ hour\}\}
\end{quote}

\paragraph{\rev{Genetics question of Figure~\ref{fig:case_study_genotype}.}}
\begin{quote}\small\color{revcolor}
Based on this information, what is Moxie's phenotype for the wool color trait?\\
A. black wool\\
B. white wool
\end{quote}

\paragraph{\rev{Retained negative from Qwen2.5-VL-3B.}}
\begin{quote}\footnotesize\ttfamily\raggedright\color{revcolor}
<think>\\
To determine Moxie's phenotype for the wool color trait, we need to understand the genetics involved. The problem states that the gene for the wool color trait (let's denote the alleles as L for white wool and l for black wool), and the l allele is the recessive allele because it is for black wool. The fact that Moxie has the homogenous (homogeneous) genotypic condition means that Moxie's genotype is either LL or ll. Specifically:\\[0.5\baselineskip]
- If Moxie has the genotype LL, this indicates that Moxie has the white allele (L) dominant over the black allele (l), because when both alleles are the same, it results in the dominant trait. Since in this problem we need to determine Moxie's phenotype, and LL represents white wool, Moxie will have white wool as it is homozygous dominant for the white allele.\\
- If Moxie's genotype is ll, this indicates that Moxie has the recessive allele for black wool. However, the problem states that Moxie is a sheep with the homogeneous genotypic condition regarding the wool color trait which means Moxie could either be homozygous dominant (LL) or homozygous recessive (ll). Since the problem specifically says it is Moxie's homogenous genotypic condition for the wool color trait, and since L is dominant over l in this scenario, the homozygous condition (LL) will result in the dominant phenotype which is white wool.\\[0.5\baselineskip]
Since Moxie has the genotype LL (assuming Moxie to be of this genotype), it will have white wool. The homozygous dominant condition (LL) results in the white phenotype.\\
</think>\\
\textbackslash{}boxed\{B\}
\end{quote}

\paragraph{Analysis.}
\rev{In all four examples, the masks remove the query-critical evidence while the responses stay fluent and task-directed.}
In the clean image \rev{of Figure~\ref{fig:case_study_schedule}}, the River Avenue row lists 9:15 AM, 10:00 AM, 10:15 AM, and 11:00 AM, so the correct waiting time after missing the 10:15 AM train is 45 minutes.
After adversarial masking, both models produce fluent but incorrect answers under the same question.
The two failures are plausible, but both reflect language-prior shortcuts rather than correct table grounding.
The 3B model reduces the problem to a simple subtraction between a nearby visible pair of times and ignores the requirement to find the next train after 10:15 AM at the same station.
The 7B model follows a different shortcut and interprets ``next'' as the next station in the table, leading it to subtract across rows instead of staying on River Avenue.
In both cases, the weakened visual evidence breaks the row-and-column alignment needed for correct reasoning, and the models fall back to familiar textual heuristics for schedule questions.
\rev{In Figure~\ref{fig:case_study_triangles}, the clean image contains 20 triangles in six colors, including two pink triangles in the upper row.
The mask hides both pink triangles and parts of several other cells.
The response counts each color and sums the counts, which is the right procedure, but it never mentions pink and reaches a total of 16 instead of 20.
In Figure~\ref{fig:case_study_burlington}, the mask covers the Burlington arrival time, and only the fragment 0 P.M. of the departure time remains visible.
The response reads the departure as 10:00 P.M., fills in an arrival of 11:00 P.M., and reports a one-hour stop.
In the clean image, the train arrives at Burlington at 12:55 P.M. and departs at 1:10 P.M., so the correct answer is 15 minutes.
In Figure~\ref{fig:case_study_genotype}, the evidence-ablated image hides the genotype ll of Moxie, while the word homozygous and the statement that black wool is recessive remain visible.
The response correctly notes that a homozygous sheep can be either LL or ll, but it then fills in the dominant genotype LL and answers white wool.
Under the clean image, Moxie has the genotype ll and therefore black wool, so this fluent response follows a textual template rather than the visual evidence.
These examples illustrate} the type of fluent but visually ungrounded negative response that \EVAN{} is designed to elicit and suppress during training.

\paragraph{\rev{Masks on blank regions.}}
\label{app:blank_masks}
\rev{In images with large blank areas, such as Figure~\ref{fig:case_study_genotype}, part of the masked patches fall on the blank background.
This happens because the patch score in Eq.~\eqref{eq:iig_patch_score} is a gradient magnitude, which measures how strongly the scaled-IIG score responds to the pixels of a patch rather than how much visual content the patch contains.
Vision transformers repurpose tokens in low-information background areas to store global image information~\citep{darcet2024vision}, and large multimodal models allocate high attention to a few visual tokens that are irrelevant to the text~\citep{kang2025see}.
Blank patches whose tokens play these roles can therefore receive high patch scores even though they show no visual content.
In content-dense images such as Figures~\ref{fig:case_study_triangles} and~\ref{fig:case_study_burlington}, the masked patches fall on the content itself, and in Figure~\ref{fig:case_study_genotype} the mask still hides the genotype ll needed to answer the question.}

\section{Limitations}

This work shares limitations common to current large multimodal models and RLVR methods.
Although benchmark improvements indicate better performance on evaluated image-text reasoning tasks, they do not fully resolve the broader challenge of reliable visual grounding in more diverse or open-ended settings.
In addition, RLVR is inherently tied to the reward design and evaluation protocol, which capture only part of multimodal reasoning quality and may not fully reflect faithfulness, robustness, or calibration.
More broadly, strong results on curated benchmarks should not be interpreted as a complete solution to the general limitations of current multimodal reasoning systems.

\section{Broader Impact}

This work studies improved visual grounding for multimodal reasoning models.
More reliable use of visual evidence may benefit applications such as education, scientific assistance, and accessibility tools.
At the same time, stronger multimodal reasoning does not remove model errors and may increase over-trust in outputs that appear well grounded.
More capable multimodal systems could also be misused in high-stakes or persuasive settings.
We therefore view this work as a step toward more reliable LMMs, but not as a guarantee of safety or correctness in deployment.